\documentclass[a4paper,11pt]{article}

\usepackage[T1]{fontenc} 
\usepackage{float} 

\makeatletter
\gdef\@fpheader{ }
\gdef\@journal{ }
\makeatother
\RequirePackage{amsmath}
\RequirePackage{amssymb}
\RequirePackage{epsfig}
\RequirePackage{graphicx}
\RequirePackage[numbers,sort&compress]{natbib}
\RequirePackage{color}
\RequirePackage[colorlinks=true
,urlcolor=blue
,anchorcolor=blue
,citecolor=blue
,filecolor=blue
,linkcolor=blue
,menucolor=blue
,pagecolor=blue
,linktocpage=true
,pdfproducer=medialab
,pdfa=true
]{hyperref}

\newif\ifnotoc\notocfalse
\newif\ifemailadd\emailaddfalse
\newif\iftoccontinuous\toccontinuousfalse
\makeatletter
\def\@subheader{\@empty}
\def\@keywords{\@empty}
\def\@abstract{\@empty}
\def\@xtum{\@empty}
\def\@dedicated{\@empty}
\def\@arxivnumber{\@empty}
\def\@collaboration{\@empty}
\def\@collaborationImg{\@empty}
\def\@proceeding{\@empty}
\def\@preprint{\@empty}

\newcommand{\subheader}[1]{\gdef\@subheader{#1}}
\newcommand{\keywords}[1]{\if!\@keywords!\gdef\@keywords{#1}\else%
\PackageWarningNoLine{\jname}{Keywords already defined.\MessageBreak Ignoring last definition.}\fi}
\renewcommand{\abstract}[1]{\gdef\@abstract{#1}}
\newcommand{\dedicated}[1]{\gdef\@dedicated{#1}}
\newcommand{\arxivnumber}[1]{\gdef\@arxivnumber{#1}}
\newcommand{\proceeding}[1]{\gdef\@proceeding{#1}}
\newcommand{\xtumfont}[1]{\textsc{#1}}
\newcommand{\correctionref}[3]{\gdef\@xtum{\xtumfont{#1} \href{#2}{#3}}}
\newcommand\jname{JHEP}

\newcommand\preprint[1]{\gdef\@preprint{\hfill #1}}

\makeatother

\newcommand\note[2][]{%
\if!#1!%
\stepcounter{footnote}\footnotetext{#2}%
\else%
{\renewcommand\thefootnote{#1}%
\footnotetext{#2}}%
\fi}

\makeatletter
\newtoks\auth@toks
\renewcommand{\author}[2][]{%
  \if!#1!%
    \auth@toks=\expandafter{\the\auth@toks#2\ }%
  \else
    \auth@toks=\expandafter{\the\auth@toks#2$^{#1}$\ }%
  \fi
}
\makeatother
\makeatletter
\newtoks\affil@toks\newif\ifaffil\affilfalse
\newcommand{\affiliation}[2][]{%
\affiltrue
  \if!#1!%
    \affil@toks=\expandafter{\the\affil@toks{\item[]#2}}%
  \else
    \affil@toks=\expandafter{\the\affil@toks{\item[$^{#1}$]#2}}%
  \fi
}
\makeatother
\makeatletter
\newtoks\email@toks\newcounter{email@counter}%
\newcommand{\emailAdd}[1]{%
\emailaddtrue%
\ifnum\theemail@counter>0\email@toks=\expandafter{\the\email@toks, \@email{#1}}%
\else\email@toks=\expandafter{\the\email@toks\@email{#1}}%
\fi\stepcounter{email@counter}}
\newcommand{\@email}[1]{\href{mailto:#1}{\tt #1}}
\makeatother

\makeatletter
\newcommand*\collaboration[1]{\gdef\@collaboration{#1}}
\newcommand*\collaborationImg[2][]{\gdef\@collaborationImg{#2}}
\makeatletter
\newcommand\afterLogoSpace{\smallskip}
\newcommand\afterSubheaderSpace{\vskip3pt plus 2pt minus 1pt}
\newcommand\afterProceedingsSpace{\vskip21pt plus0.4fil minus15pt}
\newcommand\afterTitleSpace{\vskip23pt plus0.06fil minus13pt}
\newcommand\afterRuleSpace{\vskip23pt plus0.06fil minus13pt}
\newcommand\afterCollaborationSpace{\vskip3pt plus 2pt minus 1pt}
\newcommand\afterCollaborationImgSpace{\vskip3pt plus 2pt minus 1pt}
\newcommand\afterAuthorSpace{\vskip5pt plus4pt minus4pt}
\newcommand\afterAffiliationSpace{\vskip3pt plus3pt}
\newcommand\afterEmailSpace{\vskip16pt plus9pt minus10pt\filbreak}
\newcommand\afterXtumSpace{\par\bigskip}
\newcommand\afterAbstractSpace{\vskip16pt plus9pt minus13pt}
\newcommand\afterKeywordsSpace{\vskip16pt plus9pt minus13pt}
\newcommand\afterArxivSpace{\vskip3pt plus0.01fil minus10pt}
\newcommand\afterDedicatedSpace{\vskip0pt plus0.01fil}
\newcommand\afterTocSpace{\bigskip\medskip}
\newcommand\afterTocRuleSpace{\bigskip\bigskip}
\newlength{\affiliationsSep}
\newcommand\beforetochook{\pagestyle{myplain}\pagenumbering{roman}}

\DeclareFixedFont\trfont{OT1}{phv}{b}{sc}{11}

\renewcommand\maketitle{
\pagestyle{empty}
\thispagestyle{titlepage}
\setcounter{page}{0}
\noindent{\small\scshape\@fpheader}\@preprint\par

\afterLogoSpace
\if!\@subheader!\else\noindent{\trfont{\@subheader}}\fi
\afterSubheaderSpace
\if!\@proceeding!\else\noindent{\sc\@proceeding}\fi
\afterProceedingsSpace
{\LARGE\flushleft\sffamily\bfseries\@title\par}
\afterTitleSpace
\hrule height 1.5\p@%
\afterRuleSpace
\if!\@collaboration!\else
{\Large\bfseries\sffamily\raggedright\@collaboration}\par
\afterCollaborationSpace
\fi
\if!\@collaborationImg!\else
{\normalsize\bfseries\sffamily\raggedright\@collaborationImg}\par
\afterCollaborationImgSpace
\fi
{\bfseries\raggedright\sffamily\the\auth@toks\par}
\afterAuthorSpace
\ifaffil\begin{list}{}{%
\setlength{\leftmargin}{0.28cm}%
\setlength{\labelsep}{0pt}%
\setlength{\itemsep}{\affiliationsSep}%
\setlength{\topsep}{-\parskip}}
\itshape\small%
\the\affil@toks
\end{list}\fi
\afterAffiliationSpace
\ifemailadd 
\noindent\hspace{0.28cm}\begin{minipage}[l]{.9\textwidth}
\begin{flushleft}
\textit{E-mail:} \the\email@toks
\end{flushleft}
\end{minipage}
\else 
\PackageWarningNoLine{\jname}{E-mails are missing.\MessageBreak Plese use \protect\emailAdd\space macro to provide e-mails.}
\fi
\afterEmailSpace
\if!\@xtum!\else\noindent{\@xtum}\afterXtumSpace\fi
\if!\@abstract!\else\noindent{\renewcommand\baselinestretch{.9}\textsc{Abstract:}}\ \@abstract\afterAbstractSpace\fi
\if!\@keywords!\else\noindent{\textsc{Keywords:}} \@keywords\afterKeywordsSpace\fi
\if!\@arxivnumber!\else\noindent{\textsc{ArXiv ePrint:}} \href{http://arxiv.org/abs/\@arxivnumber}{\@arxivnumber}\afterArxivSpace\fi
\if!\@dedicated!\else\vbox{\small\it\raggedleft\@dedicated}\afterDedicatedSpace\fi
\ifnotoc\else
\iftoccontinuous\else\newpage\fi
\beforetochook\hrule
\tableofcontents
\afterTocSpace
\hrule
\afterTocRuleSpace
\fi
\setcounter{footnote}{0}
\pagestyle{myplain}\pagenumbering{arabic}
} 

\renewcommand{\baselinestretch}{1.1}\normalsize
\divide\@tempdima\baselineskip
\@tempcnta=\@tempdima
\skip\@mpfootins = \skip\footins
\renewcommand{\@dotsep}{10000}

\newcommand\ps@myplain{
\pagenumbering{arabic}
\renewcommand\@oddfoot{\hfill-- \thepage\ --\hfill}
\renewcommand\@oddhead{}}
\let\ps@plain=\ps@myplain

\newcommand\ps@titlepage{\renewcommand\@oddfoot{}\renewcommand\@oddhead{}}

\numberwithin{equation}{section}

\renewcommand\section{\@startsection{section}{1}{\z@}%
                                   {-3.5ex \@plus -1.3ex \@minus -.7ex}%
                                   {2.3ex \@plus.4ex \@minus .4ex}%
                                   {\normalfont\large\bfseries}}
\renewcommand\subsection{\@startsection{subsection}{2}{\z@}%
                                   {-2.3ex\@plus -1ex \@minus -.5ex}%
                                   {1.2ex \@plus .3ex \@minus .3ex}%
                                   {\normalfont\normalsize\bfseries}}
\renewcommand\subsubsection{\@startsection{subsubsection}{3}{\z@}%
                                   {-2.3ex\@plus -1ex \@minus -.5ex}%
                                   {1ex \@plus .2ex \@minus .2ex}%
                                   {\normalfont\normalsize\bfseries}}
\renewcommand\paragraph{\@startsection{paragraph}{4}{\z@}%
                                   {1.75ex \@plus1ex \@minus.2ex}%
                                   {-1em}%
                                   {\normalfont\normalsize\bfseries}}
\renewcommand\subparagraph{\@startsection{subparagraph}{5}{\parindent}%
                                   {1.75ex \@plus1ex \@minus .2ex}%
                                   {-1em}%
                                   {\normalfont\normalsize\bfseries}}

\def\fnum@figure{\textbf{\figurename\nobreakspace\thefigure}}
\def\fnum@table{\textbf{\tablename\nobreakspace\thetable}}

\long\def\@makecaption#1#2{%
  \vskip\abovecaptionskip
  \sbox\@tempboxa{\small #1. #2}%
  \ifdim \wd\@tempboxa >\hsize
    \small #1. #2\par
  \else
    \global \@minipagefalse
    \hb@xt@\hsize{\hfil\box\@tempboxa\hfil}%
  \fi
  \vskip\belowcaptionskip}

\renewenvironment{thebibliography}[1]{%
\begin{oldthebibliography}{#1}%
\small%
\raggedright%
\setlength{\itemsep}{5pt plus 0.2ex minus 0.05ex}%
}%
{%
\end{oldthebibliography}%
}

\begin{document}


\title{\boldmath An Effective and Efficient Method  
to Solve the High-Order and the Non-Linear Ordinary 
Differential Equations: the Ratio Net}

\author[a]{Chen-Xin Qin,}
\author[b,1]{Ru-Hao Liu,}\note{Ru-Hao Liu and Chen-Xin Qin contributed equivalently to this work.}
\author[a]{Mao-Cai Li,}
\author[a,2]{Chi-Chun Zhou,}\note{zhouchichun@dali.edu.cn. Corresponding author.}
\author[a,3]{and Yi-Liu}\note{liuyi@dali.edu.cn. Corresponding author.}

\affiliation[a]{School of Engineering, Dali University, Dali, Yunnan 671003, PR China}
\affiliation[b]{School of Information Engineering, Nanchang University, Nanchang, Jiangxi 330000, PR China}









\abstract{
An effective and efficient method that solves the high-order and the non-linear ordinary 
differential equations is provided. The method is based on the ratio net.
By comparing the method with existing methods such as the 
polynomial based method and the multilayer perceptron network based method, 
we show that the ratio net gives good results and has higher efficiency.}

\maketitle
\flushbottom

\section{Introduction}

Many problems encountered in the fields of 
engineering, physics, mathematics, and e.t.c. can be described by differential equations. 
For example, differential equations are used in 
the simulation of oscillation phenomena \cite{2002Runge}, the social sciences \cite{Ledder2005Differential},
the treatment of tumors \cite{duclous2010deterministic}, and e.t.c.. 
In many cases, problems are expressed in the form of the 
higher-order and the non-linear ordinary differential equations.
For example, a high-order differential equation filter can effectively 
remove the noise in a signal \cite{2015High}. 
However, to analytically solve the ordinary differential  
equations (ODEs), especially the higher-order and the non-linear ODEs, is difficult. 
For example, the analytical methods such as 
the series method \cite{zaitsev2002handbook,corliss1982solving}
and the constant-transform method \cite{zaitsev2002handbook}
sometimes fail at the higher-order and the non-linear cases. 

Among many numerical methods, 
the neural network based methods are prominent and give good results 
universally to various kinds of ODEs. 
For example, neural network based methods are used to solve the first-order linear \cite{LEE1990110} 
and nonlinear ODEs \cite{mall2013regression}. 
S. Chakraverty and S. Mall provide a regression-based weight generation 
algorithm \cite{chakraverty2014regression} and a
Legendre neural network \cite{2016Application} to solve high-order ODEs.
YunLei Yang et al. prove the effectiveness of using neural networks to solve ODEs \cite{2018A}; 


In order to construct an effective neural network structure, 
several methods are provided and these methods can be 
divided into two categories: (1) the neural network based method with the 
trial functions based on the multilayer perceptron  
network (MLP) \cite{kumar2011multilayer,2003The,2009Multilayer,schneidereit2020solving},
the radial basis function network (RBF) \cite{2017Approximate}, and e.t.c.. 
(2) The polynomial based method with the trial function based on 
the Legendre polynomials \cite{2016Application,2018A}, the 
Chebyshev polynomials \cite{2016Single,2014Chebyshev,2020Single,juhola2008intelligent}, and e.t.c..

Beyond the construction of the trial functions, the boundary condition of ODEs 
brings other difficulties. Usually, researchers consider the boundary 
conditions as an additional constraint by transforming it into an extra term in the loss function \cite{2018A}. 
In that approach, the parameters of the model are trained to not only fit the shape of the target function 
but also meet the boundary condition. At cases, the boundary condition prevent the neural network
from finding the target function. 
Another approach introduces a trial function by constructing a network that 
automatically satisfies the boundary conditions. 
For example, a trial function in the form $y^{trial}\left(x\right)=Net\left(x\right)B\left(x\right)+g\left(x\right) $, 
where $B\left(x\right)$ and $g\left(x\right)$ are 
designed to satisfy the boundary conditions 
and $Net\left(x\right)$ are designed to search the target function \cite{2006Numerical}.

There are two key issues in solving the ODEs by using the neural network based methods, 
namely effectiveness and efficiency. 
(1) Effectiveness is about how to construct an effective neural network structure that is 
able to find the target function, or the solution 
of the ODE. (2)
Efficiency is about how to accelerate the training process.
For the first issue, researchers provide various kinds of neural network 
based methods and have proved the effectiveness.
Beyond the effectiveness, the efficiency need to be considered too.
That is, methods that can solves the ODEs 
efficiently are preferred. 

In this paper, an effective and efficient 
method that solves the high-order 
and the non-linear ODEs is provided. 
The method is based on the ratio net, which is proposed 
in the previous works \cite{zhou2020activation,zhou2020pade}.
Based on the previous work \cite{zhou2020activation,zhou2020pade}, 
we show that the ratio net is both efficiency and effective in searching the 
target function.
We solve the illustrative examples from Refs. \cite{2018A,2013AN,2016An,Ay2011The,2016Solving}.
By comparing the method with existing methods such as the 
polynomial based method and the MLP based method, 
we show that the ratio net is both efficient and effective.

This paper is organized as follows: 
in Sec. 2, we give the description of the method.
In Sec. 3, we show the advantage of the method by applying 
it on various illustrative examples. 
In Sec. 4, conclusions are given.

\section{Method description}
In this section, we introduce the main method used in the present paper, including 
the structure of ratio net, the trial function under boundary conditions, the loss function,
and the updating algorithm for the weights. 

\subsection{The ratio net: a brief review}

The method in the present paper are based on the ratio net \cite{zhou2020activation}. 
The ratio net is proved to be more efficient than the conventional neural 
networks such as the MLP
and the RBF \cite{zhou2020activation}.
In this section, we give a brief review on the ratio net.

\begin{figure}[H]
\centering
\includegraphics[width=1.0\textwidth]{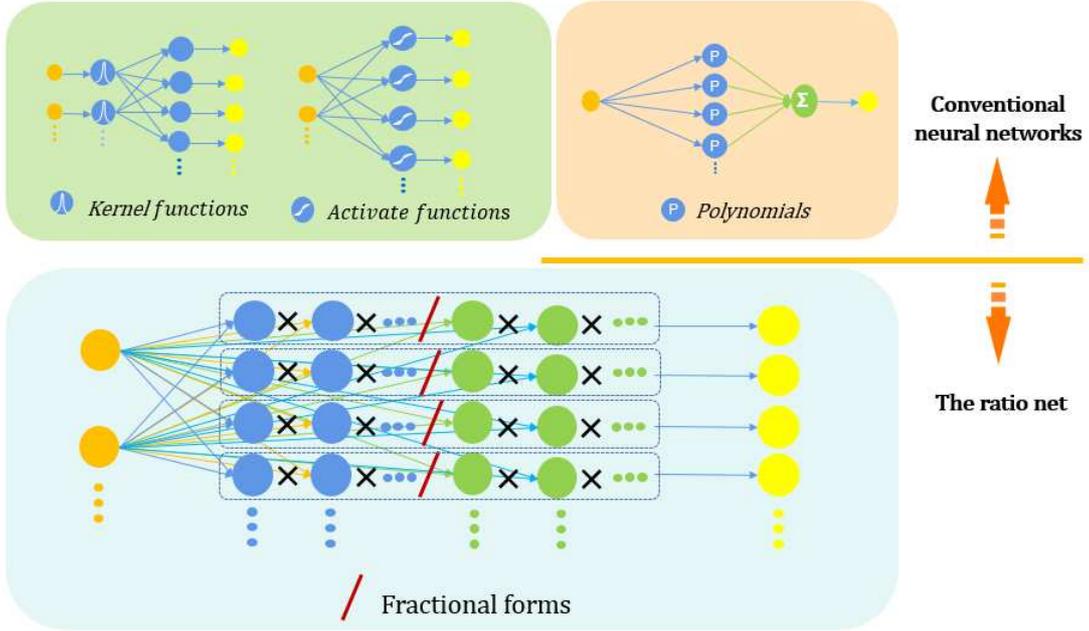}
\caption{A diagram of Ratio neural networks}
\label{Ratio neural network}
\end{figure}

A neural network is good at searching the mode of the relation between the outputs and the inputs.
To search the mode, the nonlinearity between the outputs and the inputs is the main difficulty.
Instead of the nonlinear activation function and the nonlinear kernel function, 
the ratio net uses fractional forms to solve the difficulty of nonlinearity, 
see Fig. (\ref{Ratio neural network}). 
In the ratio net, the relation between the outputs and the inputs are 
\begin{equation}
y^{Ratio}_{i}(x)=\frac{\left(\sum_{j=1}^nw^{\prime}_{ij}x_j+b^{\prime}_i\right)\left(\sum_{j=1}^nw^{\prime\prime}_{ij}x_j+b^{\prime\prime}_i\right)....}
{\left(\sum_{j=1}^nw^{\prime\prime\prime}_{ij}x_j+b^{\prime\prime\prime}_i\right)\left(\sum_{j=1}^nw^{\prime\prime\prime\prime}_{ij}x_j+b^{\prime\prime\prime\prime}_i\right)....}
\label{Ratio}
\end{equation}
Where $w$, $b$, and e.t.c. are parameters or weights. $j$ runs from $1$ to the dimension of the input $x$ 
and $i$ runs from $1$ to the dimension of the output $y$.
Here, in order to solve the ODEs, the dimension of input and output are both $1$.

\subsection{The trial function under the boundary condition}
In solving an ODE, the boundary condition brings another difficulty.
We find that to construct a network as the trial function that meet the constraint of the boundary condition 
automatically is important.
In this section, based on the ratio net, we construct a trial function that can automatically
meet the constraint of the boundary condition. 

Here, we consider ODEs in the interval $[a,b]$ with boundary conditions $y^{(n)}(a)=A_{n}$ and $y^{(m)}(b)=B_{m}$,  
where $y^{(n)}(a)$ and $y^{(m)}(b)$ are the $n$-th and $m$-th derivative at $a$ and $b$ respectively. 
$A_{n}$ and $B_{m}$ are real numbers representing the boundary values. For example, a set of boundary conditions are $y^{(0)}(a)=A_{0}$, $y^{(0)}(b)=B_{0}$,
 $y^{(2)}(a)=A_{2}$, and $y^{(2)}(b)=B_{2}$.
The trial function in the present paper is in the form:
\begin{equation}
y^{trial}_{ratio}(x)=y^{ratio}(x)(x-a)^{M_{a}}(b-x)^{M_{b}}+g(x),
\label{trial}
\end{equation} 
where $M_{a}$ and $M_{b}$ are the maximum order of derivative in the boundary condition at $a$ and $b$ respectively.
$y_{ratio}(x)$ is the ratio net giving in Eq. (\ref{Ratio}) with the dimension of input and output both $1$.
$g(x)$ is a series of highest order $L$ with $L$ the number of equations in the boundary conditions. $g(x)$ meet the boundary conditions
and thus is decided by the boundary conditions. For example, if the boundary conditions read $y^{(0)}(a)=A$ and $y^{(0)}(b)=B$,
$g(x)= (xB-xA+bA-aB)/(b-a)$.  
\begin{figure}[H]
\centering
\includegraphics[width=1.0\textwidth]{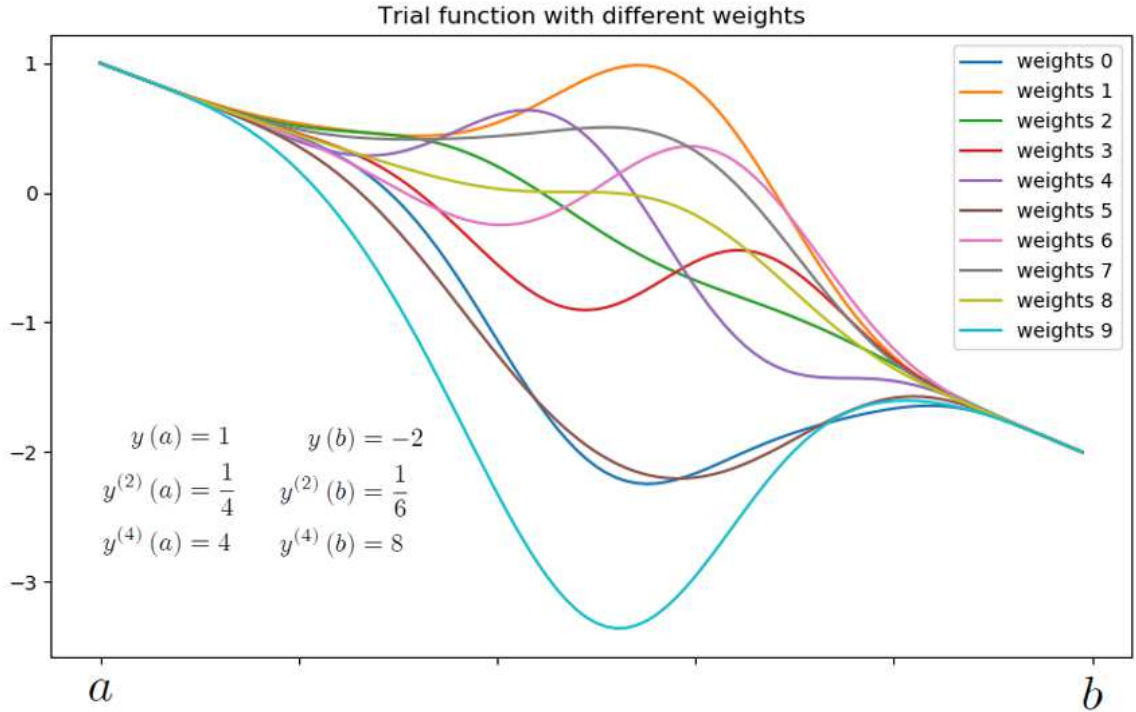}
\caption{An example of the trial function. By adjusting the weights, the trial function is capable of
searching the target function that meet the boundary conditions.}
\label{An example of the trial function}
\end{figure}

\subsection{Other existing methods: a brief review}
In order to show the efficiency of the new method in solving the ODEs, we 
compare the proposed method with other existing methods as well.
Without loss of generality, we consider two typical conventional neural networks that
are used to solve the ODEs. They are the Legendre polynomial based method and the MLP based method. In this section, we give a brief review on these methods.

\textit{The polynomial based methods: the Legendre polynomials.} The 
Legendre polynomial is a typical class of orthogonal polynomials.
The linear combination of Legendre polynomials up to order $L$ can be used 
as a effective function approximant.
The Legendre polynomials read:
\begin{equation}
P_0(x)=1,
\end{equation}
\begin{equation}
P_1(x)=x,
\end{equation} 
and 
\begin{equation}
P_{n+1}(x)=\frac{2n+1}{n+1}xP_{n}(x)-\frac{n}{n+1}P_{n-1}(x),
\end{equation} 
where $P_{l}(x)$ is Legendre polynomials of order $l$.
The trial function based on the Legendre polynomials is 
\begin{equation}
y^{trial}_{Legendre}(x)=y^{Legendre}(x)(x-a)^{M_{a}}(b-x)^{M_{b}}+N(x),
\label{legendre}
\end{equation} 
where
\begin{equation}
y^{Legendre}(x)=\sum_{j=0}^L\omega_{j}P_j(x)
\label{legendre1}
\end{equation}
with $\omega_{j}$ the weight to be decided.
 
\textit{The neural network based methods: the MLP.} 
The MLP is a classical neural network that consists of an input layer, several hidden layers, and an output layer. 
To obtain the next layer, a nonlinear activation function is applied to the linear combination of the previous layer.
For example, a one-hidden layer MLP with the dimension of input layer and output layer $1$ is given as 
\begin{equation}
y^{MLP}(x)= \sum_{j=1}^n\omega_{ij}\sigma\left(\sum_{i=1}^L\omega_{i}x +b_1 \right)+b_2,
\end{equation} 
where $L$ is the size of hidden layer and $\sigma(x) $ is the activation function.
Usually, the activation function can be 
$\tanh\left(x\right)=\left(e^x-e^{-x}\right)/\left(e^x+e^{-x}\right)$,
$sigmod\left(x\right)=1/\left(1+e^{-x}\right)$,  
and $relu\left(x\right)=x$ for $x >0$ and $0$ for $x<0$.
The trial function of neural network based methods read:
\begin{equation}
y^{trial}_{MLP}(x)=y^{MLP}(x)(x-a)^{M_{a}}(b-x)^{M_{b}}+N(x),
\label{MLP}
\end{equation} 

In this work, we consider trial functions as follows:
\[
y^{trial}=\left\{
\begin{array}
[c]{c}%
y^{ratio}\left(  x\right)  \left(  x-a\right)  ^{M_{a}}\left(  b-x\right)
^{M_{b}}+g\left(  x\right)  \\
y^{Legendre}\left(  x\right)  \left(  x-a\right)  ^{M_{a}}\left(  b-x\right)
^{M_{b}}+g\left(  x\right)  \\
y^{MLP}\left(  x\right)  \left(  x-a\right)  ^{M_{a}}\left(  b-x\right)
^{M_{b}}+g\left(  x\right)
\end{array}
\right. \label{xxxx}
\] 

\subsection{The loss function and the update of the weights}

The general form of differential equations with boundary conditions can be expressed as
\begin{equation}
F[x,y(x),y^{(1)}(x),.....,y^{(k)}(x)]=0,  x \in [a,b]
\end{equation}
and 
\begin{equation}
C[x,y(x),y^{(1)}(x),.....,y^{(l)}(x)]=0,  x=a,b,
\end{equation}
where $n$ is the order of the ODE and $C(x)$ is the boundary condition of the ODE.

Since the trial function, Eq. (\ref{trial}), meet the boundary conditions automatically,
in solving such ODEs, we only need to minimize
the following loss function 
\begin{equation}
loss=\begin{Bmatrix} F[x,y^{trial}(x),\frac{dy^{trial}}{dx},.....,\frac{d^ny^{trial}}{dx^n}] \end{Bmatrix}^2
\label{loss}
\end{equation} 
by training the ratio net.

The wrights in the ratio net are trained to minimize the loss function Eq. (\ref{loss}) 
through the gradient descent method:
\begin{equation}
\omega'=\omega + k\frac{\partial loss}{\omega}.
\end{equation}
The iteration of weight in the ratio net is implemented through tensorflow. 
The details of the implementation of the algorithm will
be given in the github for each illustrative example.

\section{Illustrative examples}
In this section, 
we apply the ratio net on various illustrative examples.
Illustrative examples from 
Refs. \cite{2018A,2016An,2013AN,Ay2011The,2016Solving} are included.
We compare the ratio net with other representative methods.
It shows that the ratio net is with higher efficiency.

\subsection{Non-linear ordinary differential equations.}   

\begin{figure}[H]
\centering
\includegraphics[width=1.0\textwidth]{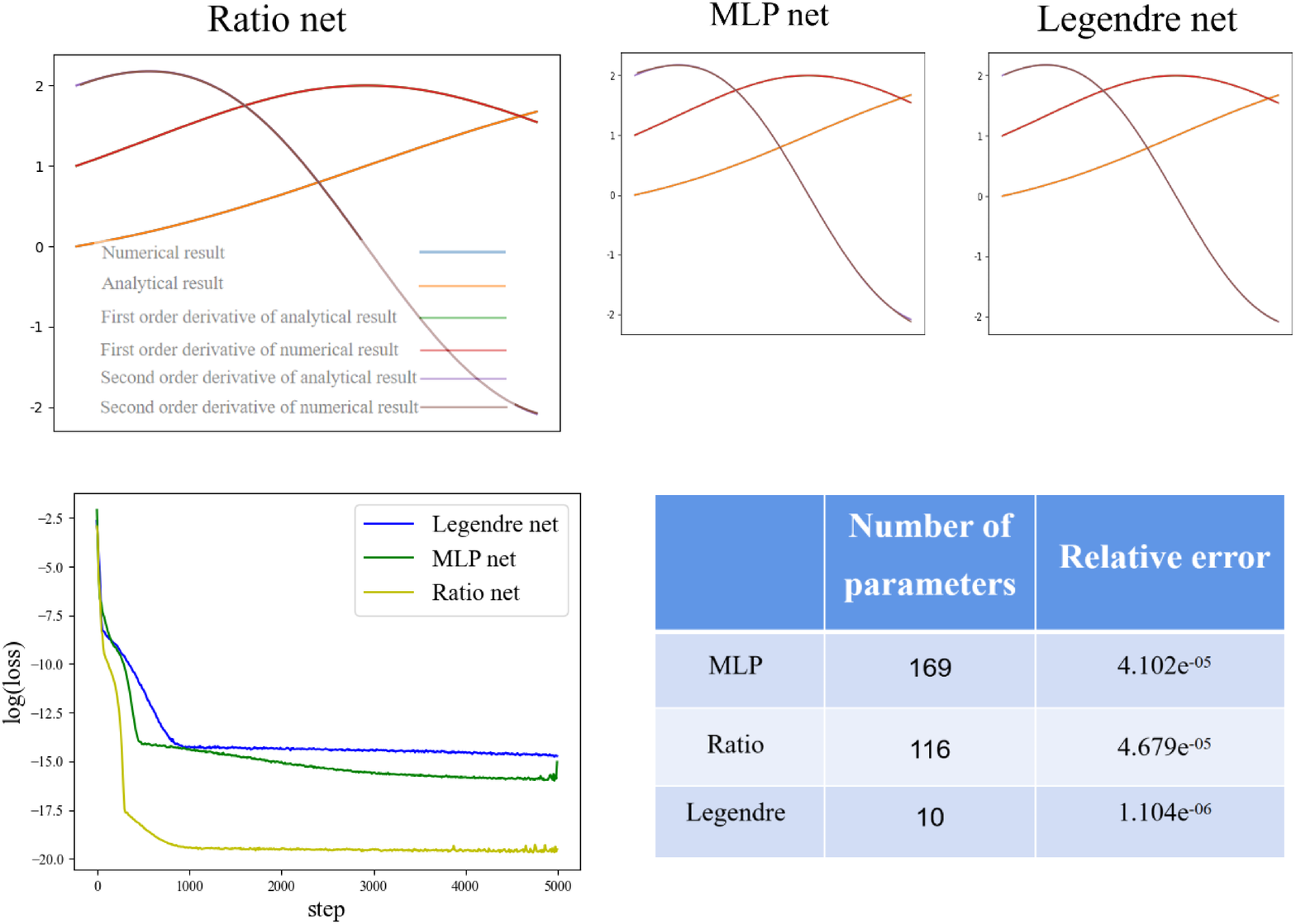}
\caption{Comparison of the results given by the three neural networks of example 1. 
The learning rates are all $0.01$.}
\label{example1}
\end{figure}

\textit{Example 1}. We consider the ODE 
\begin{equation}
y^{\prime}\left(x\right)=2y\left(x\right)-y^2\left(x\right)+1                
\end{equation}
with boundary conditions 
\begin{equation}
y\left(0\right)=1+\sqrt{2}\tanh\left[\frac{1}{2}\log\left(\frac{\sqrt{2}-1}{\sqrt{2}+1}\right)\right]                        
\end{equation}  
and 
\begin{equation}
y\left(1\right)=1+\sqrt{2}\tanh\left[\sqrt{2}+\frac{1}{2}\log\left(\frac{\sqrt{2}-1}{\sqrt{2}+1}\right)\right].                        
\end{equation} 
The exact solution of example 1 is 
\begin{equation}
y\left(x\right)=1+\sqrt{2}\tanh\left[\sqrt{2}x+\frac{1}{2}\log\left(\frac{\sqrt{2}-1}{\sqrt{2}+1}\right)\right].
\end{equation} 
Fig. (\ref{example1}) shows the results given by three neural networks.
The effectiveness is characterized by the 
fitting diagram and the relative error between the numerical solution and the analytical solution. 
The decreasing trend 
of the loss function shows the efficiency of the method.

\noindent

\begin{figure}[H]
\centering
\includegraphics[width=1.0\textwidth]{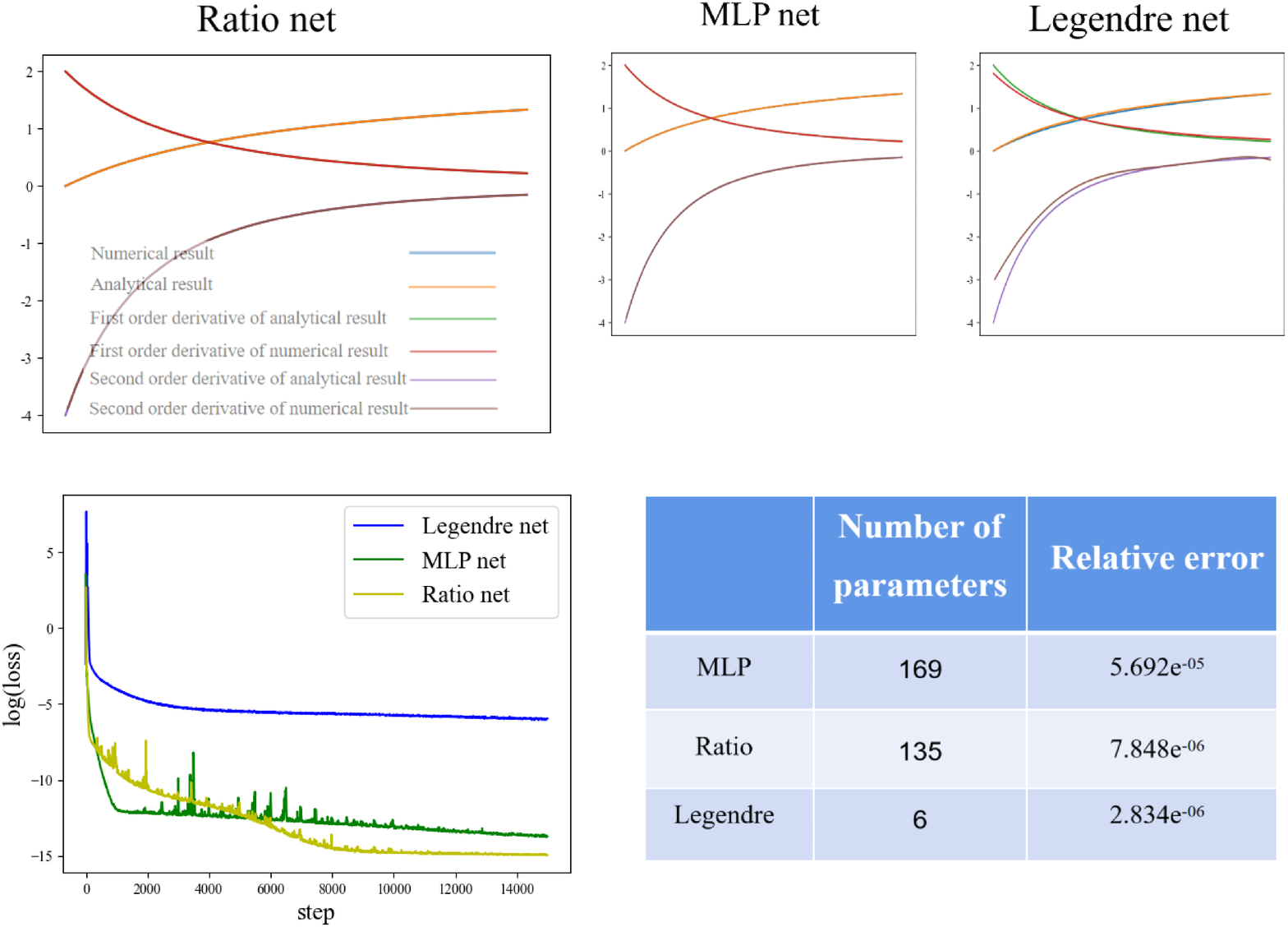}
\caption{Comparison of the results given by the three neural networks of example 2. 
The learning rates are all $0.1$.}
\label{example2}
\end{figure}

\textit{Example 2}. We consider the ODE 
\begin{equation}
y^{\prime\prime}\left(x\right)= \frac{y^3\left(x\right)-2y^2\left(x\right)}{2x^2}              
\end{equation} 
with boundary conditions 
\begin{equation}
y\left(0\right)=0\text{ and }y\left(2\right)=\frac{4}{3},
\end{equation}
which has the exact solution $ y\left(x\right)=2x/\left(x+1\right)$. Fig. (\ref{example2}) shows the results.

\noindent

\begin{figure}[H]
\centering
\includegraphics[width=1.0\textwidth]{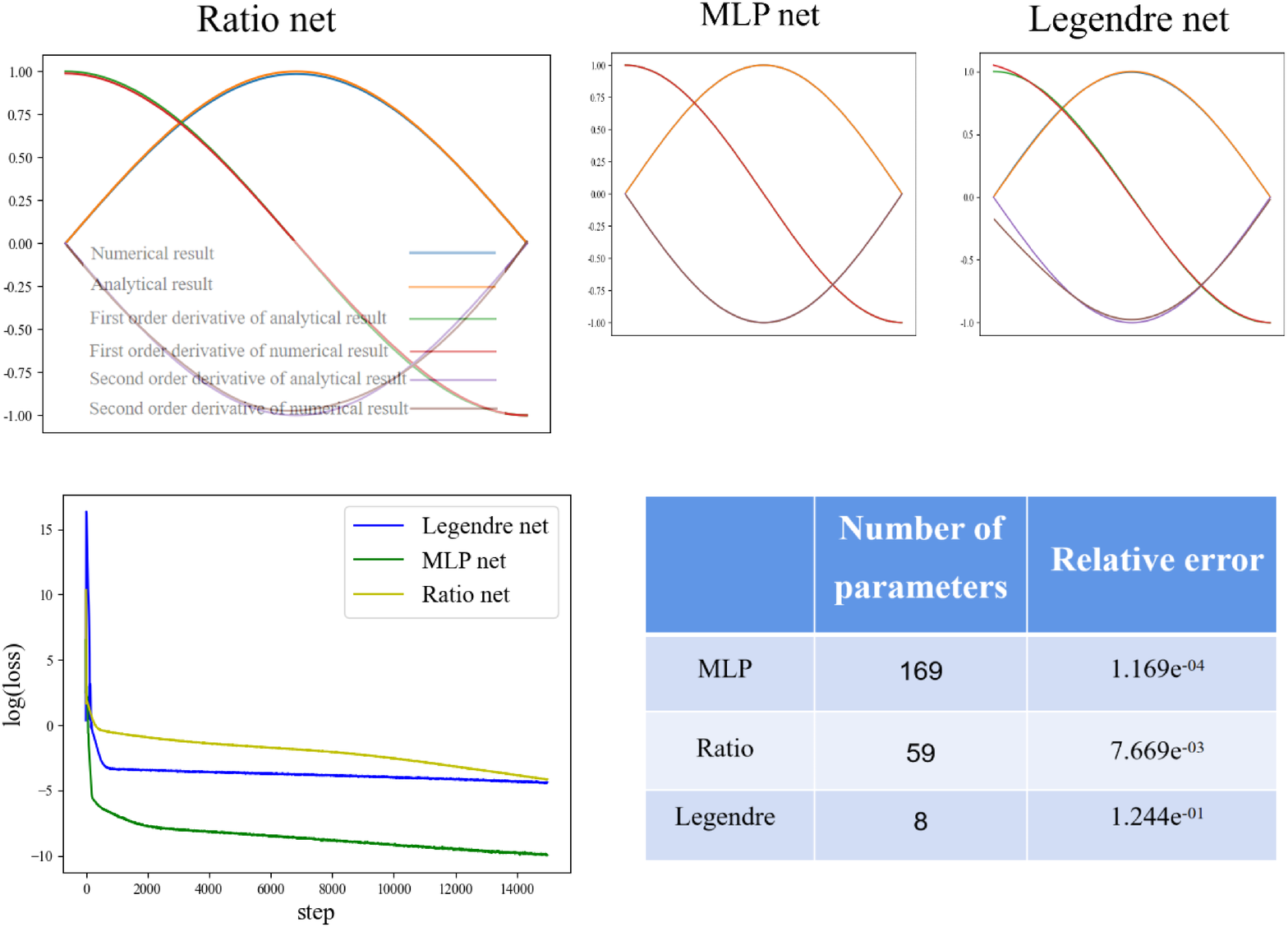}
\caption{Comparison of the results given by the three neural networks of example 3. 
The learning rates are all $0.01$.}
\label{example3}
\end{figure}

\textit{Example 3}. We consider the ODE 
\begin{equation}
y^{\prime\prime\prime}\left(x\right)=-y^2\left(x\right)-\cos\left(x\right)+\sin^2\left(x\right)                
\end{equation} 
with boundary conditions
\begin{equation}
y\left(0\right)=0\text{, }y^{\prime}\left(0\right)=1\text{, and }y\left(\pi\right)=0,
\end{equation}
which has the exact solution $ y\left(x\right)=\sin\left(x\right) $. Fig. (\ref{example3}) shows the results.

\subsection{High-order ordinary differential equations.} 

\begin{figure}[H]
\centering
\includegraphics[width=1.0\textwidth]{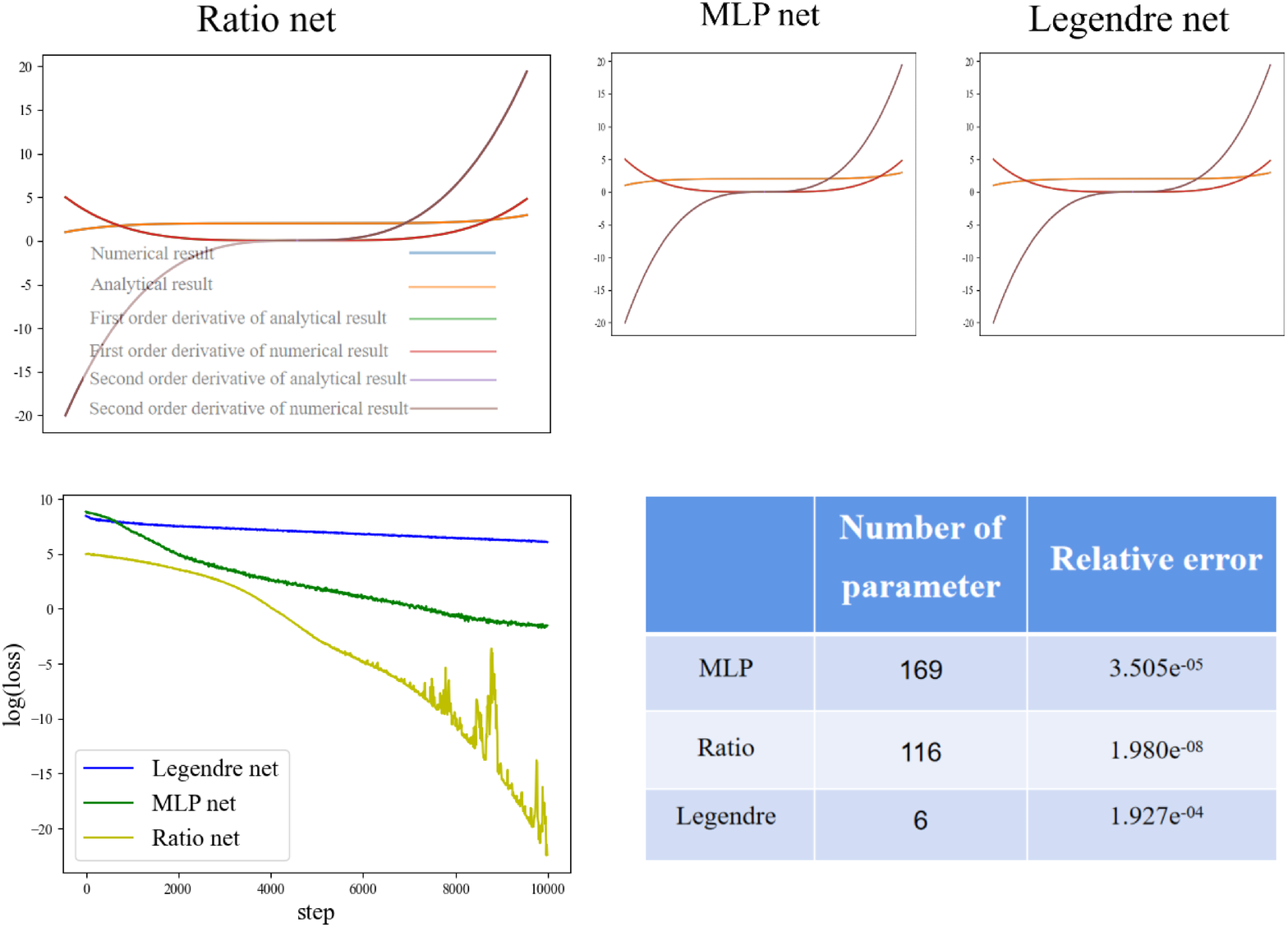}
\caption{Comparison of the results given by the three neural networks of example 4. 
The learning rates are all $0.0001$.}
\label{example4}
\end{figure}

\textit{Example 4}. We consider the ODE 
\begin{equation}
y^{\left(4\right)}\left(x\right)=120x                
\end{equation} 
with boundary conditions
\begin{equation}
y\left(-1\right)=1\text{, } y^{\prime}\left(-1\right)=5\text{, } y\left(1\right)=3\text{, and } y^{\prime}\left(1\right)=5,
\end{equation} 
which has the exact solution $ y\left(x\right)=x^5+2 $. Fig. (\ref{example4}) shows the results.

\noindent

\begin{figure}[H]
\centering
\includegraphics[width=1.0\textwidth]{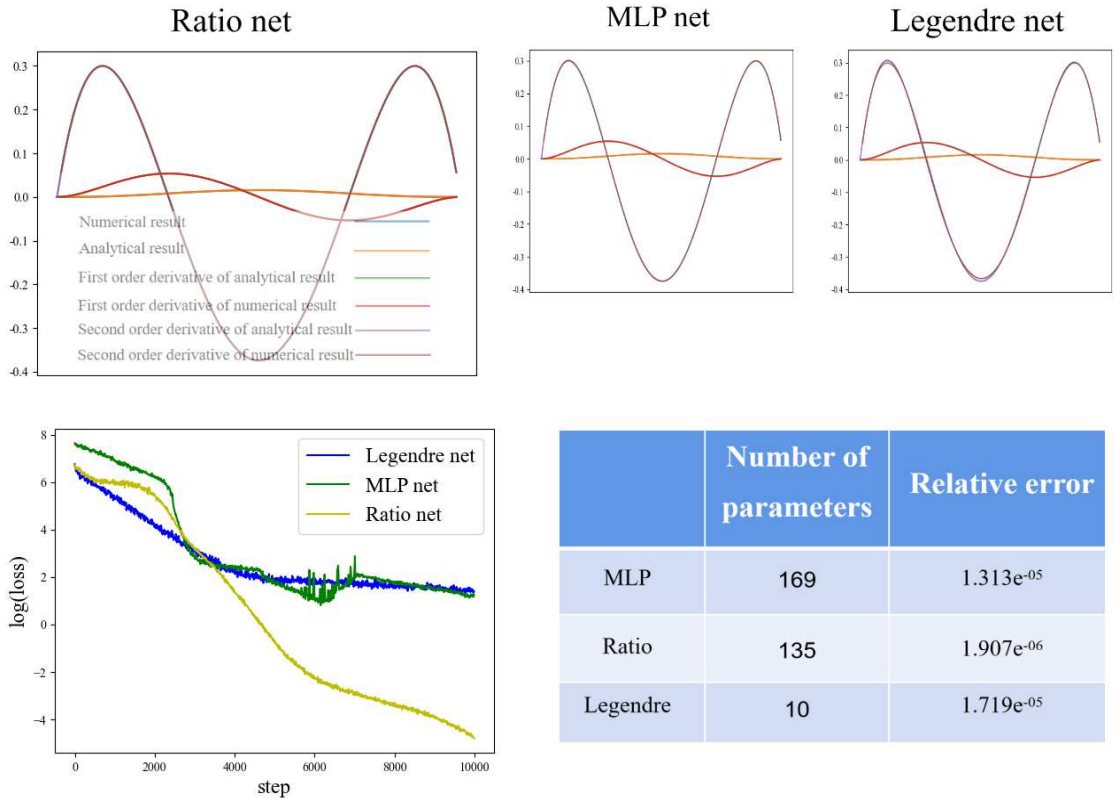}
\caption{Comparison of the results given by the three neural networks of example 5. 
The learning rates are all $0.0001$. }
\label{example5}
\end{figure}

\textit{Example 5}. We consider the ODE 
\begin{equation}
y^{(4)}\left(x\right)=-\frac{x^2}{1+y^2\left(x\right)}-72\left(1-5x+5x^2\right)+\frac{x^2}{1+{\left(x-x^2\right)}^6}                
\end{equation} 
with boundary conditions
\begin{equation}
y\left(0\right)=0\text{, } y^{\prime}\left(0\right)=0\text{, } y\left(1\right)=0\text{, and } y^{\prime}\left(1\right)=0, 
\end{equation} 
which has the exact solution $ y\left(x\right)=x^3{\left(1-x\right)}^3 $. Fig. (\ref{example5}) shows the results.

\noindent

\begin{figure}[H]
\centering
\includegraphics[width=1.0\textwidth]{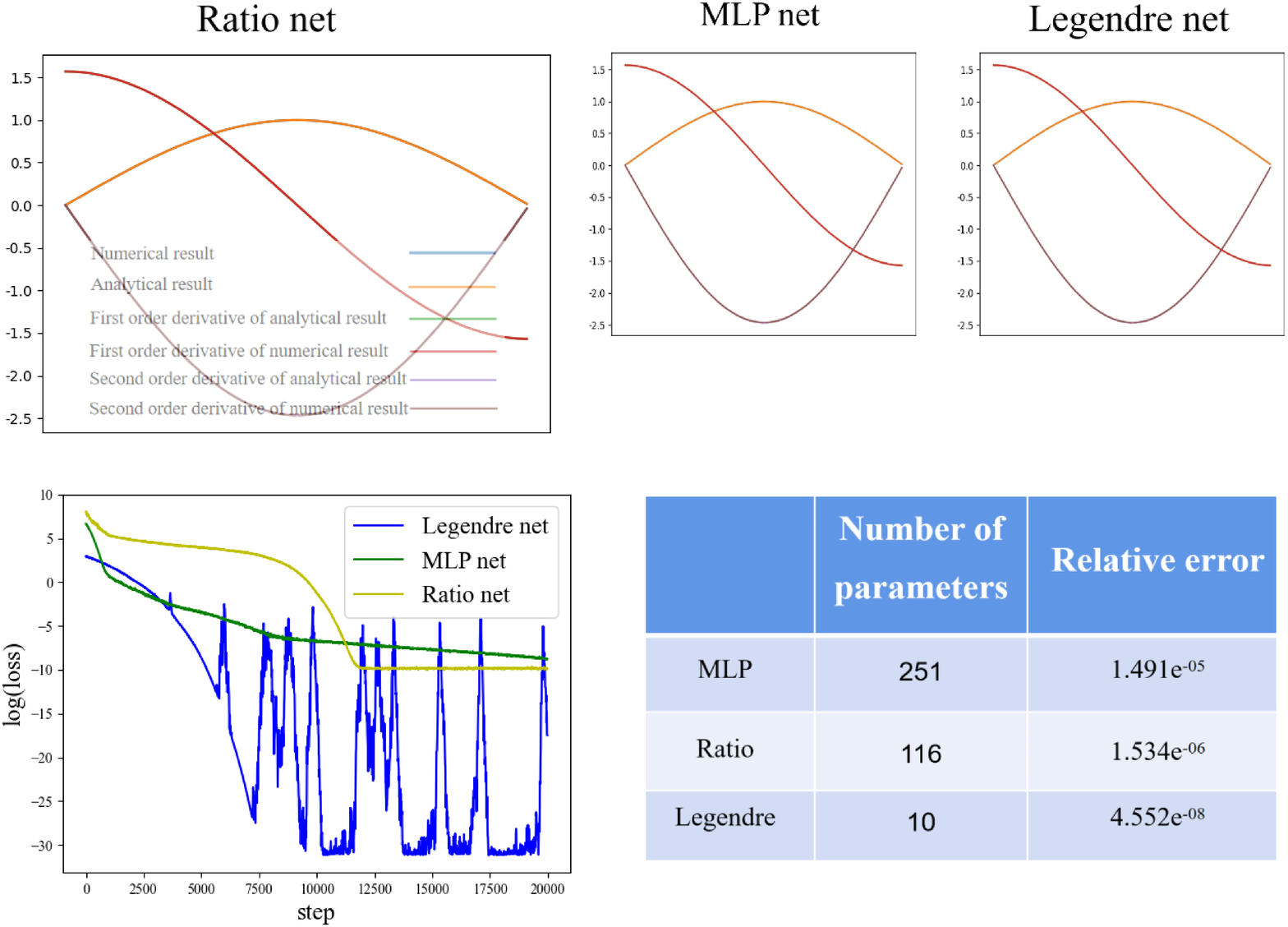}
\caption{Comparison of the results given by the three neural networks of example 6. 
The learning rates are all $0.0001$.}
\label{example6}
\end{figure}

\textit{Example 6}. We consider the ODE 
\begin{equation}
y^{(4)}\left(x\right)+y\left(x\right)=\left[{\left(\frac{\pi}{2}\right)}^4+1\right]\cos\left(\frac{\pi}{2}x\right)              
\end{equation} 
with boundary conditions
\begin{equation}
y\left(-1\right)=0 \text{, } y^{\prime}\left(-1\right)=\frac{\pi}{2}\text{, } y\left(1\right)=0\text{, and } y^{\prime}\left(1\right)=-\frac{\pi}{2},      
\end{equation} 
which has the exact solution $ y\left(x\right)=\cos\left(\pi \frac{x}{2}\right) $. Fig. (\ref{example6}) shows the results. 

\noindent

\begin{figure}[H]
\centering
\includegraphics[width=1.0\textwidth]{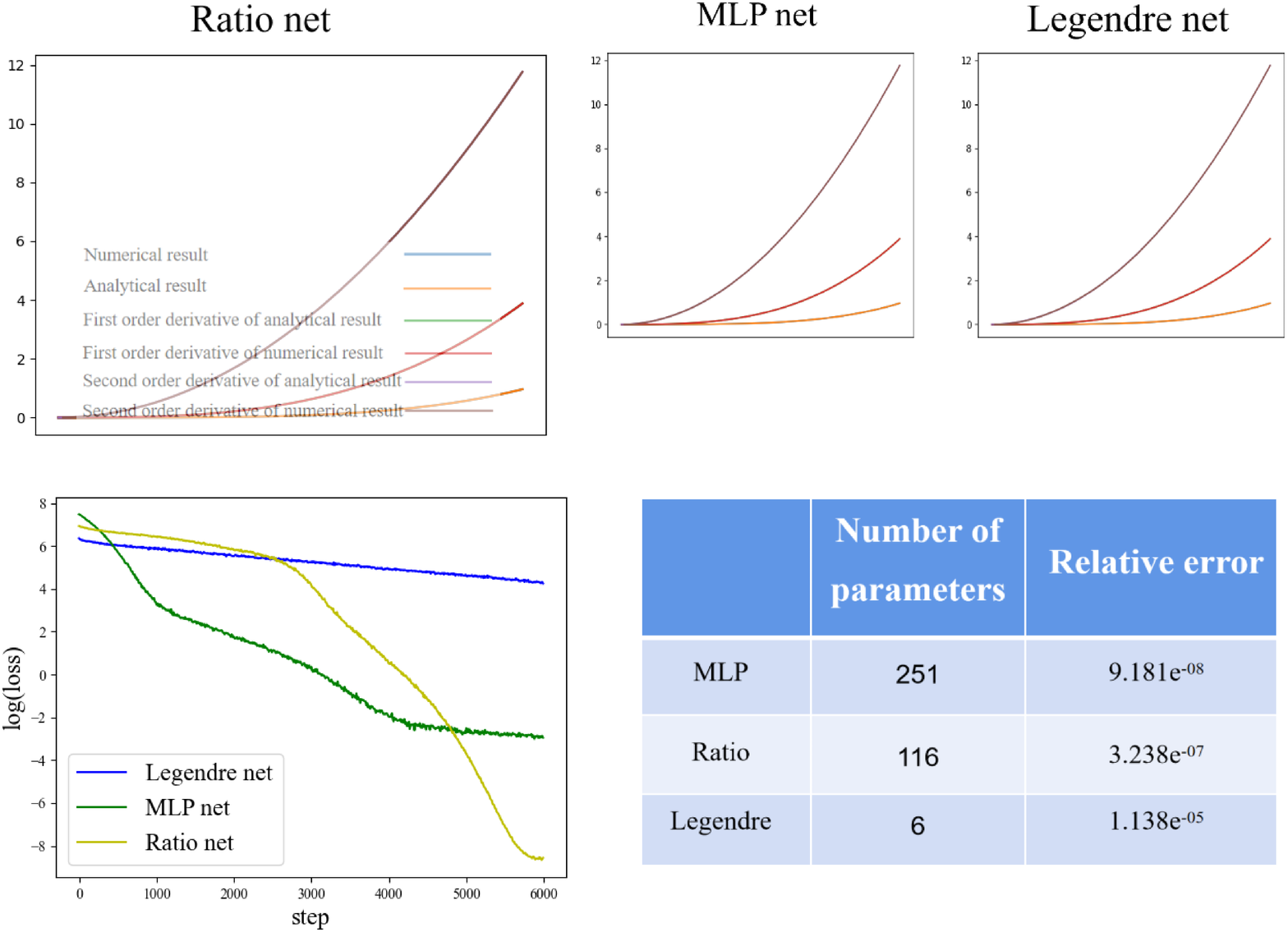}
\caption{Comparison of the results given by the three neural networks of example 7. 
The learning rates are all $0.0001$.}
\label{example7}
\end{figure}

\textit{Example 7}. We consider the ODE

\begin{equation}
y^{(4)}\left(x\right)+y^{\prime}\left(x\right)=4x^{3}+24                
\end{equation} 
with boundary conditions 
\begin{equation}
y\left(0\right)=0\text{, } y^{\prime}\left(0\right)=0\text{, } y\left(1\right)=1\text{, and }y^{\prime}\left(1\right)=4,      
\end{equation} 
which has the exact solution $ y\left(x\right)=x^4 $. Fig. (\ref{example7}) shows the results. 

\subsection{Non-linear and high-order ordinary differential equations.}

\begin{figure}[H]
\centering
\includegraphics[width=1.0\textwidth]{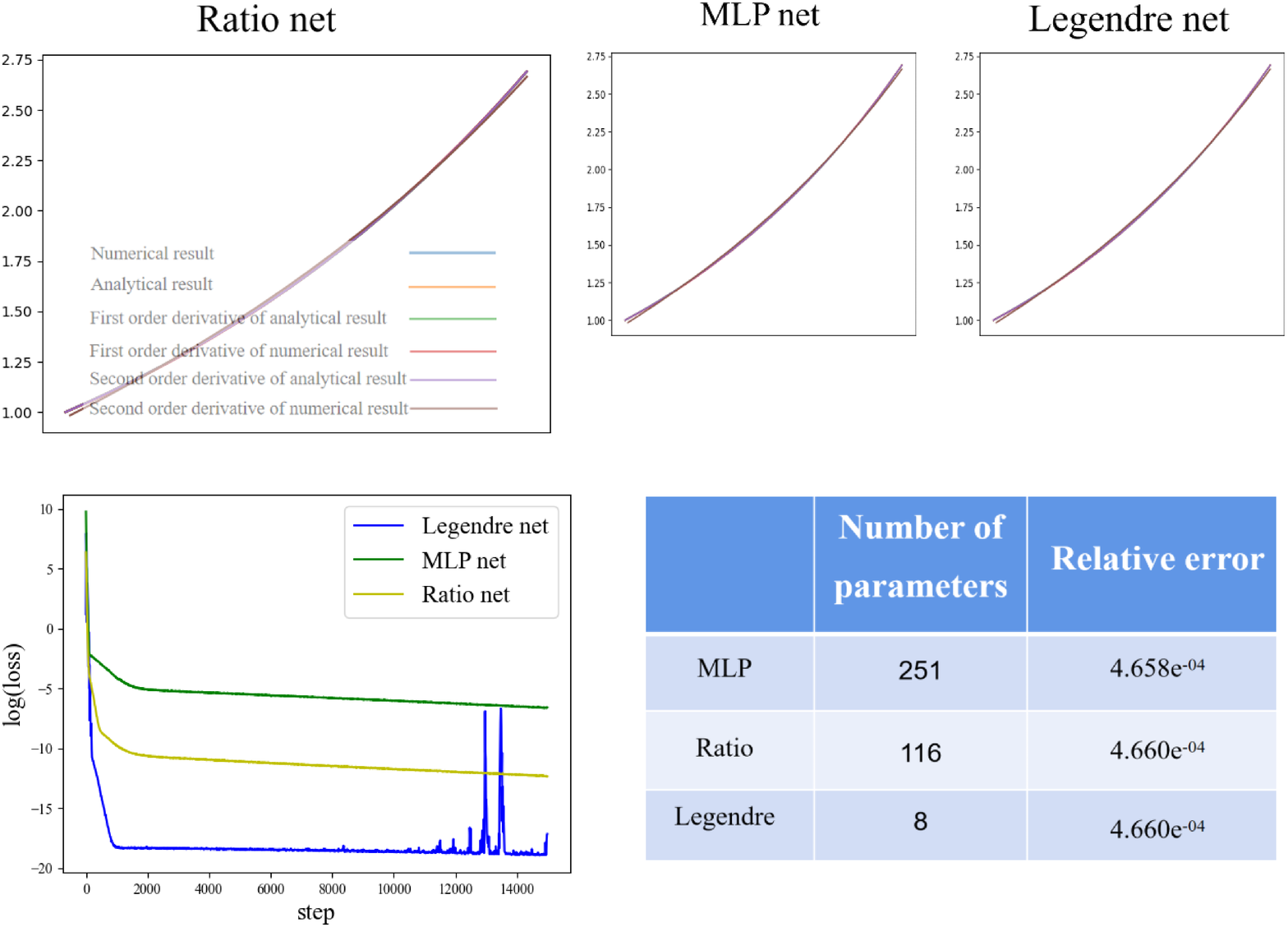}
\caption{Comparison of the results given by the three neural networks of example 8. 
The learning rates are all $0.01$.}
\label{example8}
\end{figure}

\textit{Example 8}. We consider the ODE

\begin{equation}
y^{(5)}\left(x\right)=e^{-x}y^2\left(x\right)              
\end{equation} 
with boundary conditions 
\begin{equation}
y\left(0\right)=1\text{, } y^{\prime}\left(0\right)=1\text{, } y^{\prime\prime}\left(0\right)=1\text{, } y\left(1\right)=e\text{, and }y^{\prime}\left(1\right)=e ,       
\end{equation} 
which has the exact solution $ y\left(x\right)=e^x $. Fig. (\ref{example8}) shows the results. 

\noindent

\begin{figure}[H]
\centering
\includegraphics[width=1.0\textwidth]{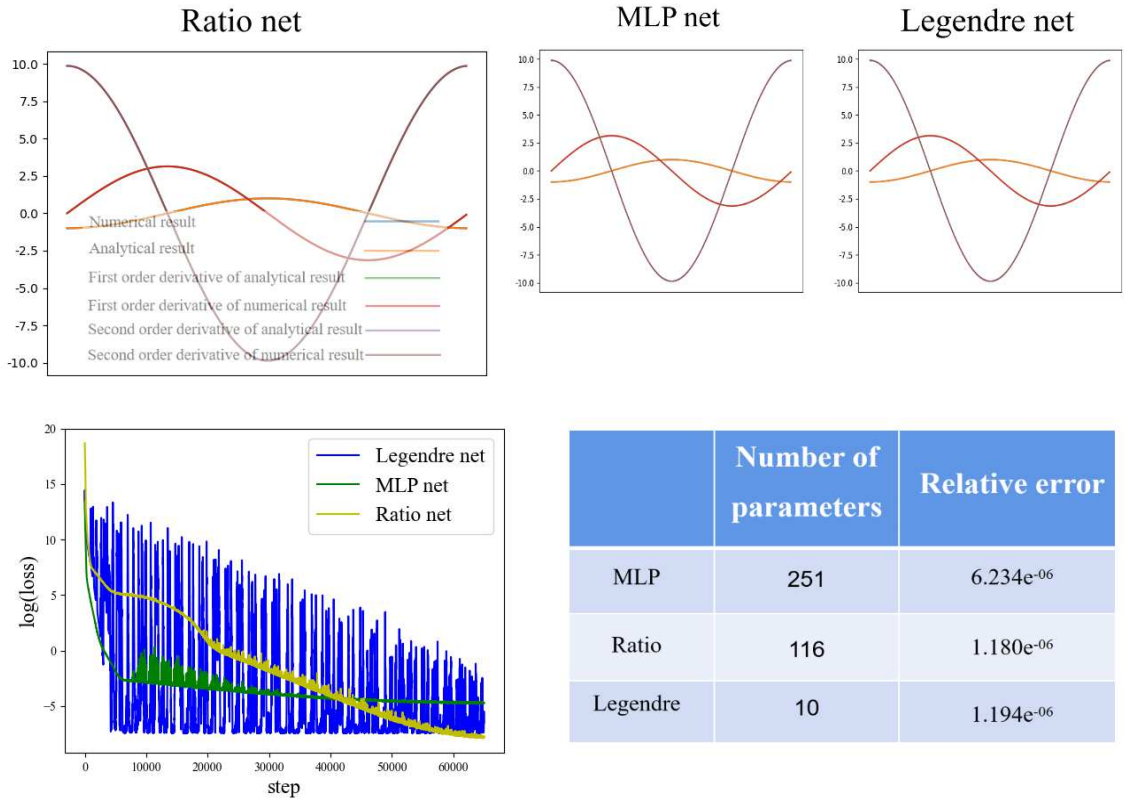}
\caption{Comparison of the results given by the three neural networks of example 9. 
The learning rates are all $0.001$.}
\label{example9}
\end{figure}

\textit{Example 9}. We consider the ODE

\begin{equation}
y^{(6)}\left(x\right)+y\left(x\right)y^{\prime\prime}\left(x\right)+y^{\prime}\left(x\right)y^{(5)}\left(x\right)-\pi^2\sin\left(\pi x\right)y^{\prime\prime\prime}\left(x\right)+\pi^1y^2\left(x\right)=-\pi^6\cos\left(\pi x\right)                
\end{equation} 
with boundary conditions
\begin{equation}
y\left(-1\right)=\cos\left(-\pi\right)\text{, } y^{\prime}\left(-1\right)=-\pi \sin\left(-\pi\right)\text{, } y^{\prime\prime}\left(-1\right)=-\pi^2\cos\left(-\pi\right),                     
\end{equation}
\begin{equation}
y\left(1\right)=\cos\left(\pi\right)\text{, } y^{\prime}\left(1\right)=-\pi\sin\left(\pi\right)\text{, and } y^{\prime\prime}\left(1\right)=-\pi^2\cos\left(\pi\right),                 
\end{equation}  
which has the exact solution $ y\left(x\right)=\cos\left(\pi x\right) $. Fig. (\ref{example9}) shows the results. 

\subsection{Linear ordinary differential equations.} 
In this section, we consider several linear ODEs.

\textit{Example 10},
\begin{equation}
y^{\prime\prime}\left(x\right)-y^{\prime}\left(x\right)=-2\sin\left(x\right)                      
\end{equation} 
with boundary conditions
\begin{equation}
y\left(0\right)=-1\text{ and }y\left(\frac{\pi}{2}\right)=1 ,      
\end{equation} 
which has the exact solution $ y\left(x\right)=\sin\left(x\right)-\cos\left(x\right)$.  
 
\textit{Example 11}, 
\begin{equation}
y^{\prime}\left(x\right)+\frac{1}{5}y=e^{-\frac{x}{5}}\cos\left(x\right)                         
\end{equation} 
with boundary conditions
\begin{equation}
y\left(0\right)=0 \text{ and } y\left(1\right)=\frac{\sin\left(1\right)}{e^{1/5}} ,     
\end{equation} 
which has the exact solution $ y\left(x\right)=e^{-x/5}\sin\left(x\right) $.  
 
\textit{Example 12},
\begin{equation}
y^{\prime}\left(x\right)+\left(x+\frac{1+3x^2}{1+x+x^3}\right)y\left(x\right)=x^3+2x+\frac{x^2+3x^4}{1+x+x^3}                        
\end{equation}
with boundary conditions
\begin{equation}
y\left(0\right)=-1\text{ and }y\left(1\right)=1+\frac{1}{3e^{1/2}} ,       
\end{equation}
which has the exact solution $ y(x)=e^{-x^2/2}/\left(1+x+x^3\right) +x^2 $. 
 
\textit{Example 13}, 
\begin{equation}
y^{\prime}\left(x\right)-\sin\left(x\right)y\left(x\right)=2x-x^2\sin\left(x\right)                         
\end{equation} 
with boundary conditions
\begin{equation}
y\left(-1\right)=1 \text{ and } y\left(1\right)=1 ,       
\end{equation} 
which has the exact solution $ y\left(x\right)=x^2 $.  
 
\textit{Example 14}, 
\begin{equation}
y^{\prime\prime}\left(x\right)+xy^{\prime}\left(x\right)-4y\left(x\right)=12x^2-3x                         
\end{equation} 
with boundary conditions
\begin{equation}
y\left(0\right)=0 \text{ and } y\left(2\right)=18,       
\end{equation} 
which has the exact solution $ y\left(x\right)=x^4+x $. 
 
\textit{Example 15}, 
\begin{equation}
y^{\prime\prime}\left(x\right)-y^{\prime}\left(x\right)=-2\sin\left(x\right)                      
\end{equation} 
with boundary conditions
\begin{equation}
y\left(0\right)=-1 \text{ and } y\left(\frac{\pi}{2}\right)=1 ,      
\end{equation} 
which has the exact solution $ y\left(x\right)=\sin\left(x\right)-\cos\left(x\right) $. 
 
\textit{Example 16}, 
\begin{equation}
y^{\prime\prime}\left(x\right)+2y^{\prime}\left(x\right)+y\left(x\right)=x^2+3x+1                      
\end{equation} 
with boundary conditions
\begin{equation}
y\left(0\right)=0 \text{ and } y\left(1\right)=-e^{-1} + 1 ,       
\end{equation} 
which has the exact solution $ y\left(x\right)=-e^{-x}+x^2-x+1 $. 
 
\textit{Example 17}, 
\begin{equation}
y^{\prime\prime}\left(x\right)+\frac{1}{5}y^{\prime}\left(x\right)+y\left(x\right)=-\frac{1}{5}e^{-\frac{x}{5}}\cos\left(x\right)               
\end{equation} 
with boundary conditions
\begin{equation}
y\left(0\right)=0 \text{ and } y\left(2\right)=\sin\left(2\right)e^{-\frac{2}{5}} ,       
\end{equation} 
which has the exact solution $ y\left(x\right)=\sin\left(x\right)e^{-x/5} $. 
 
\begin{figure}[H]
\centering
\includegraphics[width=1.0\textwidth]{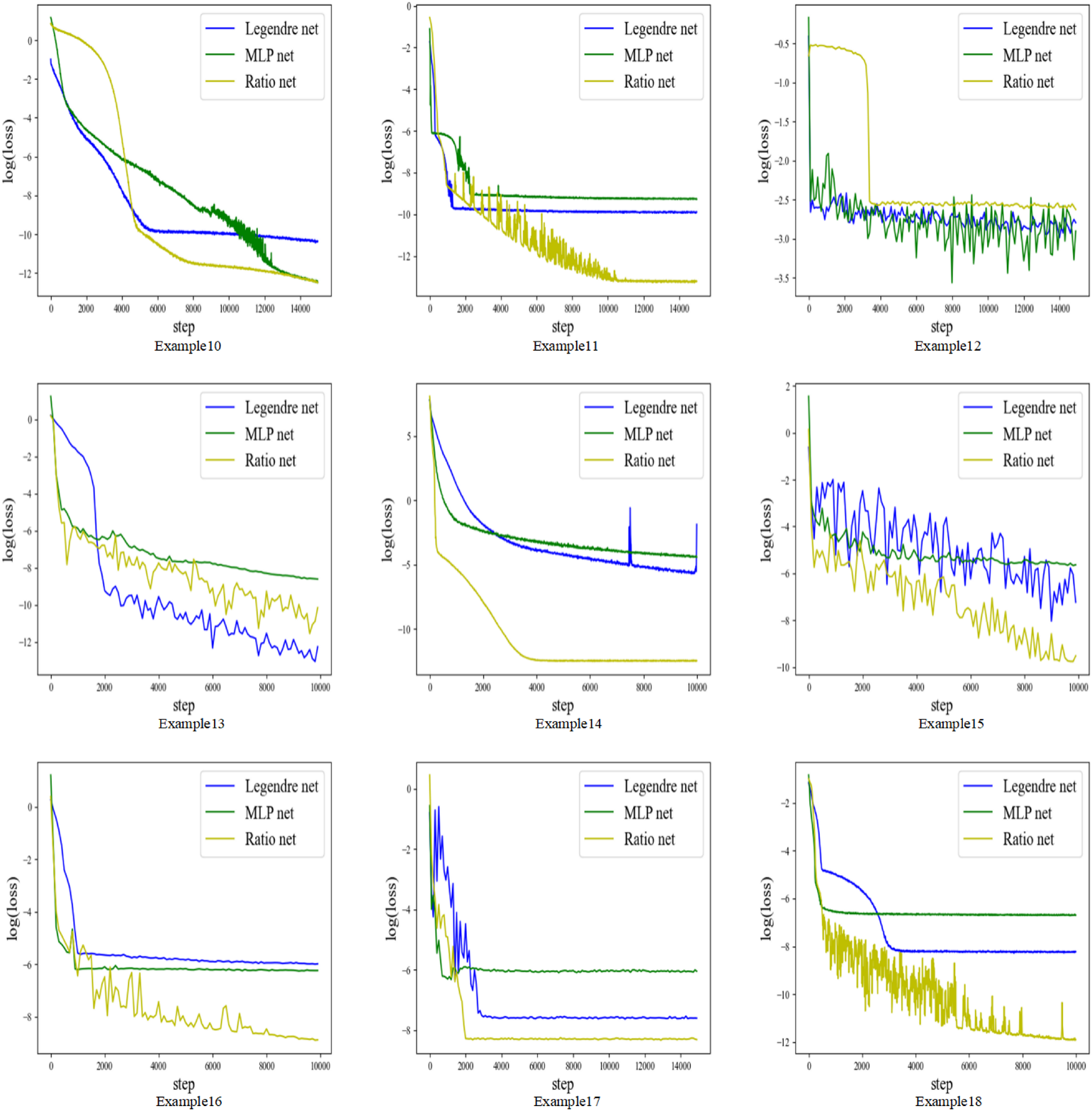}
\caption{The loss versus steps of examples 10 to 18.
The learning rates of each examples may be different, but for the same examples, the learning rates of different models are same.}
\label{example1018}
\end{figure}

\textit{Example 18},
\begin{equation}
y^{\prime}\left(x\right)=y\left(x\right)-x^2+1                     
\end{equation} 
with boundary conditions
\begin{equation}
y\left(0\right)=0.5 \text{ and } y\left(1\right)=4 - \frac{\pi}{2} ,       
\end{equation} 
which has the exact solution $ y\left(x\right)={\left(x+1\right)}^2-0.5e^x $. 
Fig. (\ref{example1018}) shows the results of examples 11-18 
given by the three neural networks.
The effectiveness is characterized by the 
fitting diagram and the relative error between the analytical solution 
and the numerical solution. 
The decreasing trend 
of the loss function shows the efficiency of the method.

\begin{table}[H]                                                           
\centering
\caption{The relative errors of examples 10–18}
\begin{tabular}{ccccccc}    
\hline
Relative error &Ratio & MLP &Legendre &  \\ \hline
Example 10  &$3.119e^{-06} $&$ 3.496e^{-04} $&$2.061e^{-05} $&  \\ \hline
Example 11  &$1.592e^{-07} $&$ 1.247e^{-06} $&$1.202e^{-06} $&  \\ \hline
Example 12  &$1.799e^{-02} $&$ 2.715e^{-02} $&$1.832e^{-02} $&  \\ \hline
Example 13  &$5.591e^{-06} $&$ 2.236e^{-05} $&$3.136e^{-07} $&  \\ \hline
Example 14  &$1.066e^{-05} $&$ 4.515e^{-06} $&$9.550e^{-05} $&  \\ \hline
Example 15  &$5.041e^{-07} $&$ 9.846e^{-04} $&$9.964e^{-05} $&  \\ \hline
Example 16  &$4.967e^{-06} $&$ 6.803e^{-06} $&$2.391e^{-06} $&  \\ \hline
Example 17  &$3.863e^{-06} $&$ 8.451e^{-03} $&$7.467e^{-06} $&  \\ \hline
Example 18  &$8.191e^{-07} $&$ 2.684e^{-05} $&$1.055e^{-05} $&  \\ \hline
\end{tabular}
\end{table}
 
\textit{Example 19},
\begin{equation}
y^{\prime\prime}\left(x\right)+y\left(x\right)=2               
\end{equation} 
with boundary conditions
\begin{equation}
y\left(0\right)=1 \text{ and } y\left(1\right)=0 ,       
\end{equation} 
which has the exact solution $ y\left(x\right)=\left[(\cos\left(1\right)-2)/(\sin\left(1\right))\right]\sin\left(x\right)-\cos\left(x\right)+2 $.
 
\textit{Example 20}, 
\begin{equation}
y^{\prime\prime}\left(x\right)+\frac{2}{x}y^{\prime}\left(x\right)+2y\left(x\right)=0               
\end{equation} 
with boundary conditions
\begin{equation}
y\left(0.001\right)=1 \text{ and } y\left(1\right)=\frac{\sin\left(\sqrt{2}\right)}{\sqrt{2}} ,      
\end{equation} 
which has the exact solution $ y\left(x\right)=\sin\left(\sqrt{2}x\right)/\sqrt{2}x $.
 
\textit{Example 21}, 
\begin{equation}
y^{\prime}\left(x\right)+\frac{\cos\left(x\right)}{\sin\left(x\right)}y\left(x\right)=\frac{1}{\sin\left(x\right)}               
\end{equation} 
with boundary conditions
\begin{equation}
y\left(1\right)=\frac{3}{\sin\left(1\right)} \text{ and } y\left(2\right)=\frac{4}{\sin\left(2\right)} ,      
\end{equation} 
which has the exact solution $ y\left(x\right)=\left(x+2\right)/\sin\left(x\right) $.
 
\textit{Example 22}, 
\begin{equation}
y^{\prime\prime}\left(x\right)-\frac{1}{1+e^x}y^{\prime}\left(x\right)-\frac{15e^{2x}}{{\left(1+e^x\right)}^2}y\left(x\right)=\frac{e^{2x}}{{\left(1+e^x\right)}^6}             
\end{equation} 
with boundary conditions
\begin{equation}
y\left(-1\right)=\frac{1}{{\left(1+e^{-1}\right)}^4} \text{ and } y\left(0\right)=\frac{1}{2^4},       
\end{equation} 
which has the exact solution $ y\left(x\right)={\left(1+e^x\right)}^{-4} $.
 
\textit{Example 23}, 
\begin{equation}
y^{\prime}\left(x\right)=4x^3-3x^2+2             
\end{equation} 
with boundary conditions
\begin{equation}
y\left(0\right)=0 \text{ and } y\left(1\right)=2 ,        
\end{equation} 
which has the exact solution $ y\left(x\right)=x^4-x^3+2x $.
 
\textit{Example 24}, 
\begin{equation}
y^{\prime}\left(x\right)=y\left(x\right)                 
\end{equation} 
with boundary conditions
\begin{equation}
y\left(0\right)=1 \text{ and } y\left(1\right)=e ,        
\end{equation} 
which has the exact solution $ y\left(x\right)=e^x $.
 
\textit{Example 25},
\begin{equation}
y^{\prime}\left(x\right)=3\cos\left(x\right)\sin^2\left(x\right)+6\sin\left(3x\right)                 
\end{equation} 
with boundary conditions
\begin{equation}
y\left(0\right)=-3 \text{ and } y\left(1\right)=-2\cos\left(3\right) + \sin^3\left(1\right) - 1,        
\end{equation} 
which has the exact solution $ y\left(x\right)=-2\cos\left(3x\right)+\sin^3x-1 $
 
\begin{figure}[H]
\centering
\includegraphics[width=1.0\textwidth]{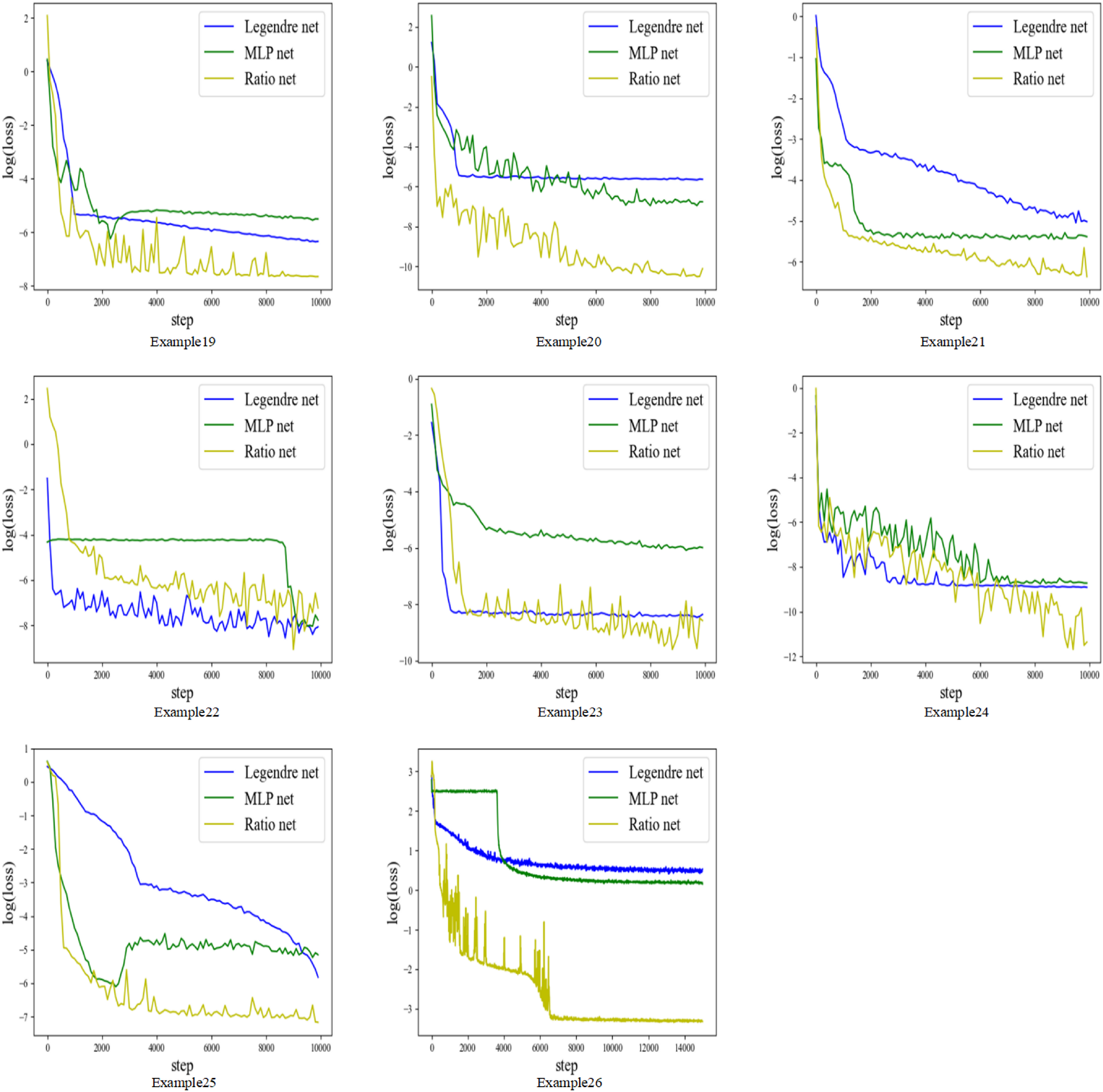}
\caption{The loss versus steps of examples 19 to 26. The learning rates of each examples may be different, but for the same examples, the learning rates of different models are same.}
\label{example1926}
\end{figure}

\textit{Example 26},
\begin{equation}
y^{\prime\prime}\left(x\right)=\frac{3}{x}\left(1-y^{\prime}\left(x\right)\right)-\frac{27}{x^3}                 
\end{equation} 
with boundary conditions
\begin{equation}
y\left(1\right)=2 \text{ and } y\left(2\right)=\frac{27}{4},       
\end{equation} 
which has the exact solution $ y\left(x\right)=x-3+27/x-23/x^2 $.
Fig. (\ref{example1926}) shows the results of examples 19-26.

\begin{table}[H]                                                           
\centering
\caption{The relative errors of examples 19–26}
\begin{tabular}{ccccccc}    
\hline
Relative error &Ratio & MLP &Legendre &  \\ \hline
Example 19  &$2.625e^{-05} $&$ 4.880e^{-05} $&$2.376e^{-06} $&  \\ \hline
Example 20  &$2.092e^{-07} $&$ 5.598e^{-05} $&$6.479e^{-06} $&  \\ \hline
Example 21  &$4.484e^{-05} $&$ 1.578e^{-04} $&$3.319e^{-04} $&  \\ \hline
Example 22  &$4.916e^{-05} $&$ 1.053e^{-05} $&$1.497e^{-05} $&  \\ \hline
Example 23  &$2.534e^{-05} $&$ 5.515e^{-05} $&$9.775e^{-06} $&  \\ \hline
Example 24  &$1.008e^{-05} $&$ 1.264e^{-07} $&$6.382e^{-06} $&  \\ \hline
Example 25  &$5.541e^{-05} $&$ 1.110e^{-04} $&$2.172e^{-04} $&  \\ \hline
Example 26  &$2.730e^{-04} $&$ 1.473e^{-02} $&$1.742e^{-02} $&  \\ \hline
\end{tabular}
\end{table}



\section{Conclusions}
In this paper, an effective and efficient 
method that solves the high-order and the non-linear ordinary 
differential equations is provided. 
The method is based on the ratio net.
By comparing the method with existing methods such as the 
polynomial based method and the MLP based method, 
we show that the ratio net is both effective and efficient.
In the following research, we use the ratio net to solve partial differential equations. 

\section{Acknowledgments}
We are very indebted to Prof. Wu-Sheng Dai for his enlightenment and encouragement. 
We are very indebted to Prof. Guan-Wen Fang and Yong-Xie for their encouragements. 
This work is supported by Yunnan Youth Basic Research Projects (202001AU070020) 
and Doctoral Programs of Dali University (KYBS201910).










\end{document}